\documentclass{article}
\usepackage{ijcai17}
\usepackage{graphicx}
\usepackage{amsmath}
\usepackage{multirow}
\usepackage{subfigure}
\usepackage[small]{caption}
\usepackage{times}

\title{Image Matching via Loopy RNN}
\author{Donghao Luo, Bingbing Ni, Yichao Yan, Xiaokang Yang\\
Shanghai Jiao Tong University\\
\{luo-donghao, nibingbing, yanyichao, xkyang\}@sjtu.edu.cn}

\begin{document}

\maketitle

\begin{abstract}
  Most existing matching algorithms are \emph{one-off} algorithms, i.e., they usually measure the distance between the two image feature representation vectors for only one time. In contrast, human's vision system achieves this task, i.e., image matching, by recursively looking at specific/related parts of both images and then making the final judgement. Towards this end, we propose a novel loopy recurrent neural network (Loopy RNN), which is capable of aggregating relationship information of two input images in a progressive/iterative manner and outputting the consolidated matching score in the final iteration. A Loopy RNN features two uniqueness. First, built on conventional long short-term memory (LSTM) nodes, it links the output gate of the tail node to the input gate of the head node, thus it brings up symmetry property required for matching. Second, a monotonous loss designed for the proposed network guarantees increasing confidence during the recursive matching process. Extensive experiments on several image matching benchmarks demonstrate the great potential of the proposed method.
\end{abstract}

\section{Introduction}
Image matching is a very important research topic in computer vision, due to its great potential in a wide range of real-world tasks including object/place retrieval~\cite{arandjelovic2016netvlad}, person re-identification~\cite{yan2016person}, $3D$ reconstruction~\cite{cheng2014fast}, etc. Mathematically, a matching algorithm takes two images as inputs and outputs a score measuring the similarity of the two inputs, i.e., higher score indicates higher similarity between the two inputs.
Previous research work is mainly focused on two aspects. On one hand, various image patch descriptors such as SIFT~\cite{lowe2004distinctive}, SURF~\cite{bay2006surf}, ORB~\cite{rublee2011orb}, etc., have been proposed to well represent the two patches, based on which the computed distance (e.g., Euclidean distance) can accurately reflect the true relationship between them.
On the other hand, metric learning based methods~\cite{jia2011heavy,jain2012metric} have been developed to achieve more discriminative distance measure, which is superior to conventional Euclidean distance.
{
\begin{figure}[t]
\centering
\includegraphics[width=1\linewidth]{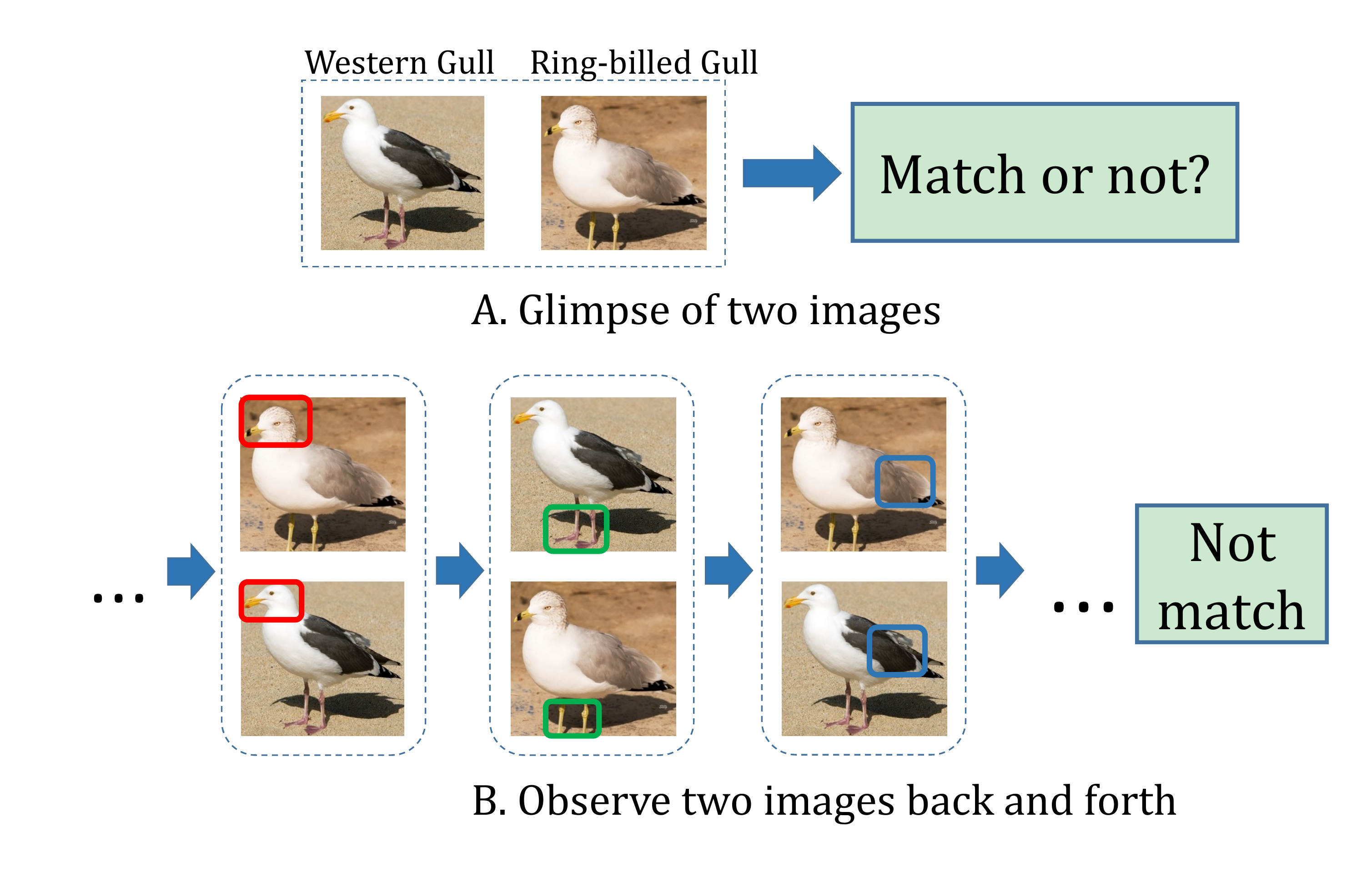}
\caption{Difference between \emph{one-off} matching and recursive matching.}
\label{fig1}
\vspace{-5mm}
\end{figure}
}

Recently, deep learning has further made significant progress in image matching on both aspects. For feature representation, SIFT based patch descriptors have been replaced with convolutional neural networks (CNN) based ones~\cite{fischer2014descriptor,paulin2015local}. The results show a significant performance gain.
For distance metric learning, end-to-end learning infrastructure has been utilized to enhance image matching. One remarkable example is the Siamese network~\cite{bromley1993signature}, in which two image patches are first input to a two-stream convolutional sub-network (with identical parameters) to extract features, and then combined with a second sub-network (based on fully connected layers) to infer the similarity of two image patches. Siamese network has been widely used in many aspects of computer vision including people re-identification~\cite{yi2014deep} and tracking~\cite{bertinetto2016fully}. Based on the end-to-end learnable capability provided by Siamese, MatchNet~\cite{han2015matchnet}~\cite{zagoruyko2015learning} has recently boosted patch-based image matching performance.

Despite their remarkable improvements, previous work on matching can all be regarded as a \emph{one-off} solution, i.e., most algorithms perform the image patch feature extraction and distance calculation for only one time and output the final matching score. However, human's vision system performs matching process in a rather \emph{recursive/iterative} manner. To judge whether the two images refer to the same species, human's attention is constantly switched between two images and moved to different patches/parts on the images. In other words, one will take turns observing different regions of both images to progressively aggregate information on the matched/un-matched portions of both images and get more and more confident. This process repeats until one is confident enough to make the final judgement. As shown in Figure~\ref{fig1}, it is difficult to distinguish if the two
birds are from same subspecies by observing the two images once. In this situation, human alternately observes two images and each observation is focused on some parts of the birds, such as the head
and leg, etc., and finally makes a confident decision based on integral and local information. From a computational point of view, the above recursive/iterative matching mechanism also has advantages over
the conventional \emph{one-off} approach, as it can progressively attend to more and more discriminative regions of the images and get rid of the issue of cluttered background or irrelevant and noisy image features.

It is thus demanding to develop a computational model or network structure to simulate the recursive mechanism to enhance image matching.
Towards this end, it is natural to consider recurrent neural networks (e.g., RNN~\cite{dorffner1996neural}, LSTM~\cite{hochreiter1997long}). Intuitively, human's attentive regions/patches on both images could be considered as a sequence of observations. This observation sequence could naturally serves as the input sequence to a RNN/LSTM structure and the aggregated similarity measure could be output from the last (temporal) node of the recurrent network. However, such a sequential model cannot be directly applied for image matching as it violates the symmetric property which is required for a valid matching algorithm. Namely, the output similarity measurement should be unchanged if we switch the order of the two input image patches. To satisfy this symmetric requirement, we propose a loopy recurrent neural network (Loopy RNN). A Loopy RNN inherits basic components and structure of conventional RNN. The major differences between a Loopy RNN structure with a conventional one are that: 1) instead of having an arbitrary number of temporal nodes, it only has two, which correspond to the two input image patches, and 2) these two nodes are cyclicly linked, thus it brings up symmetry property. When applied to image patch matching, Loopy RNN can simulate the iterative process of examining image features from both images alternatively and progressively gather more and more matched information to consolidate the final matching score.
To facilitate model training and testing, an approximation from the Loopy RNN structure toward a normal RNN/LSTM structure is developed via duplicating the head and tail nodes for a number of times. To simulate human's perception as well as to guarantee robust matching, it also requires that the confidence of similarity measurement increases when we goes deeper in our recursive matching network. For such a purpose, we utilize a monotonous objective function~\cite{ma2016learning}, which enforces more penalty to the output associated with deeper node in the network.
The proposed Loopy RNN has been experimented on several image matching benchmark including UBS patch dataset~\cite{winder2009picking} and Mikolajczyk dataset, and results demonstrate performance gain over Siamese-like networks.

\section{Related Work}
Two key components are included in image matching, one is extracting proper features from the original image and the other is measuring the distance of the features to describe the
similarity of images. At first, hand-craft features such as SIFT~\cite{lowe2004distinctive} and DAISY~\cite{tola2008fast} are cooperated with fixed metric method like Graph Model~\cite{yan2016multi,yan2015consistency,yan2015discrete}
to match images. It means that the two parts of matching (extracting feature and measuring similarity) are independent when using above methods and the isolation hinders the improvement
of performance. To break the isolation, researchers propose learning descriptors or similarity metric in condition of fixed the other part. For example,~\cite{brown2011discriminative} learns
the descriptors by minimizing the classification error and~\cite{jain2012metric} learns the metric by treating it as a linear transformation. Learning descriptors and metric jointly is proposed to make the cooperation of the two parts more powerful. In~\cite{trzcinski2012learning}, boosting trick is
adopted to learn descriptors and metrics and achieved great performance. The performance of these methods are limited by the hand-crafted features, while the proposed method employs deep features and achieves better performance.

The advent of CNNs has tremendously promoted the development of many branches of computer vision including image
matching. In~\cite{krizhevsky2012imagenet}, the performance of convolutional descriptor from AlexNet (trained on ImageNet) has been proved more effective than SIFT in most cases.
In~\cite{ren2017unsupervised}, CNN also was used to compute dense correspondence. Combining
CNN and Siamese structure~\cite{bromley1993signature}, it is natural to train the network in an end-to-end manner, i.e., learn descriptor and metric jointly. MatchNet of~\cite{han2015matchnet}
employs a Siamese network in which some convolutional layers are adopted as a feature extractor and fully connected layers as a comparator to measure similarity.
~\cite{zagoruyko2015learning} explores different architectures to do patch-based image matching including Siamese (share parameter of CNN), Pseudo-Siamese (unshare parameter of CNN),
and 2-channel (treat two patches as two channels of an image). In virtue of CNN's advantage, these methods obtain great promotion compared with previous traditional methods.
These methods mentioned above can all be regarded as \emph{one-off}, i.e., descriptors are compared just once. The Loopy RNN proposed in this paper
learns the descriptors and metric jointly like Siamese, however, draws the conclusion by repeating comparing the descriptors. Note that Shyam et al. published a paper based on similar idea, which they call Attentive Recurrent Comparators~\cite{shyam2017attentive}. We recommend the readers to also refer to this contemporary work.

\section{Methodology}
{
\begin{figure}[t]
\centering
\includegraphics[width=0.88\linewidth]{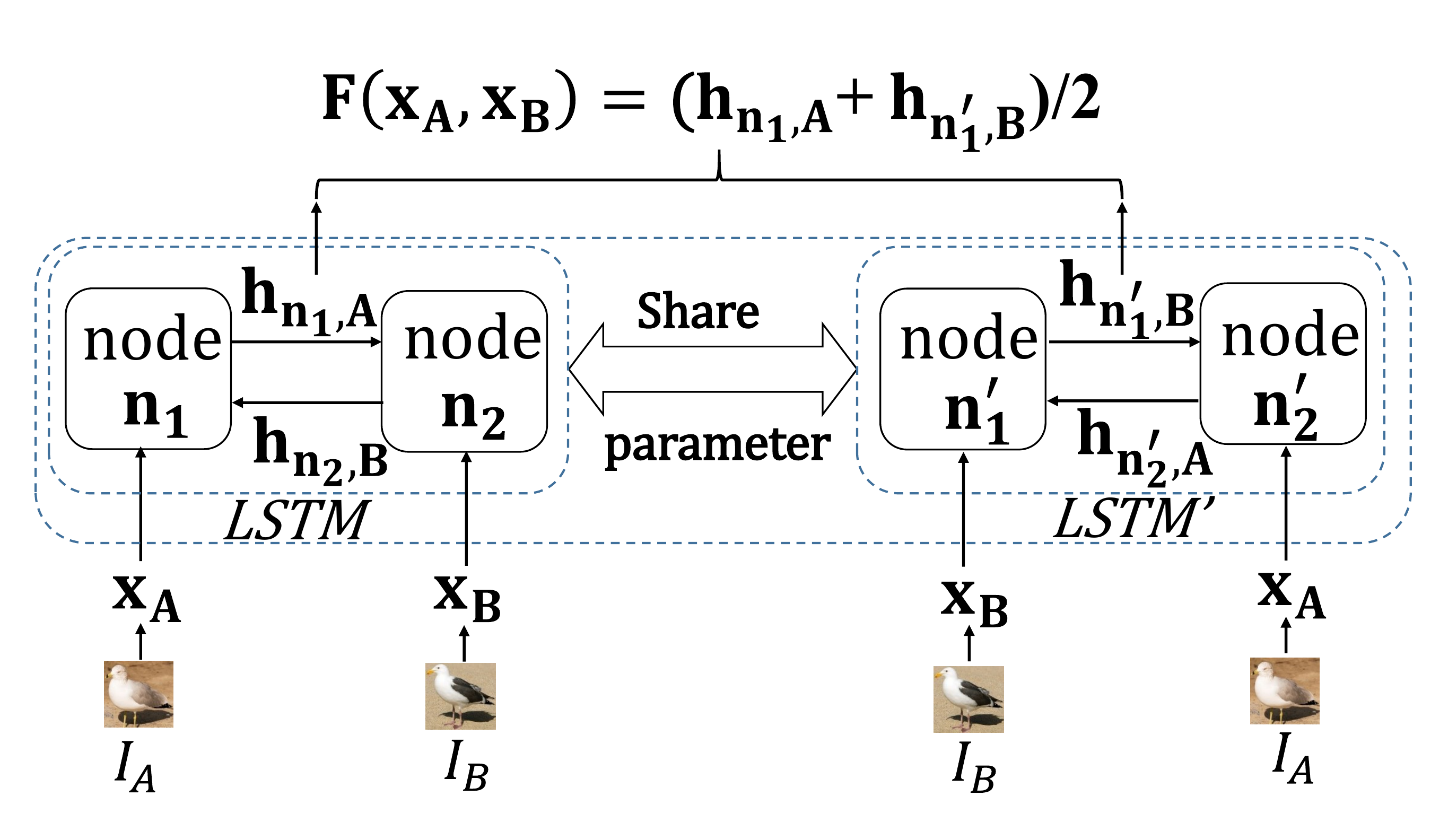}
\caption{A diagram of the proposed Loopy RNN (two nodes structure).}
\label{fig2}
\vspace{-5mm}
\end{figure}
}
The goal of this work is to develop a recusive/iterative matching framework to imitate the matching mechanism of human perception. To this end,
we propose a loopy recurrent neural network (Loopy RNN) which not only aggregates individual matching attempts and progressively yields more and more confident matching result but also preserves the symmetric property.

\subsection{Loopy Recursive Neural Network}
\textbf{Architecture}. The basic structure of a loopy recurrent neural network (Loopy RNN) is illustrated in Figure~\ref{fig2}. Our Loopy RNN consists of two sub-networks sharing parameter. Two recurrent nodes compose a sub-network. In this work, we adopt a standard long short-term memory (LSTM)~\cite{hochreiter1997long} node with input/output/forget/hidden cells as an atomic node of a Loopy RNN. Therefore we denote two
sub-networks as $LSTM$ and $LSTM^{'}$ respectively. For $LSTM$, two nodes are $\mathbf{n_1}$, $\mathbf{n_2}$. For $LSTM^{'}$, two nodes are
$\mathbf{n_1^{'}}$, $\mathbf{n_2^{'}}$. As illustrated in Figure~\ref{fig2}, the difference between a sub-network RNN and a normal RNN architecture lies in the connection between nodes, i.e., a normal RNN is a linear structure, a Loopy RNN is a circular structure with the output of the tail node linked to its head node.

We denote $\mathbf{x_{A},x_{B}}$ as a pair of inputs, i.e., features of image patches. $\mathbf{h_{A(B)}}$, $\mathbf{o_{A(B)}}$ are the hidden state and output node corresponding to
input image A (or B). For each node of LSTM, three gates input gate $\mathbf{i}$, output gate $\mathbf{o}$, and forget gate $\mathbf{f}$ as well as a memory cell
$\mathbf{c}$ are included. The LSTM nodes in our loopy network are updated as follows:
\begin{equation}
\begin{aligned}
\mathbf{i_{A(B)}}=&\sigma \mathbf{(W_{i}x_{A(B)}+U_{i}h_{B(A)}+V_{i}c_{B(A)}+b_{i})},\\
\mathbf{f_{A(B)}}=&\sigma \mathbf{(W_{f}x_{A(B)}+U_{f}h_{B(A)}+V_{f}c_{B(A)}+b_{f})},\\
\mathbf{o_{A(B)}}=&\sigma \mathbf{(W_{o}x_{A(B)}+U_{o}h_{B(A)}+V_{o}c_{A(B)}+b_{o})},\\
\mathbf{c_{A(B)}}=&\mathbf{f_{A(B)}}\odot \mathbf{c_{B(A)}}+\mathbf{i_{A(B)}}\odot \tanh\mathbf{(W_{c}x_{A(B)}}+\\
&\mathbf{U_{c}h_{B(A)}+b_{c})},\\
\mathbf{h_{A(B)}}=&\mathbf{o_{A(B)}}\odot \tanh(\mathbf{c_{A(B)}}),
\end{aligned}\label{hehe}
\end{equation}
where $\sigma$ is the sigmoid function and $\odot$ denotes the element-wise multiplication operator. $\mathbf{W_{*}}$, $\mathbf{U_{*}}$ and $\mathbf{V_{*}}$ are the weight
matrices, and $\mathbf{b_{*}}$ are the bias vectors. The memory cell $\mathbf{c_{A(B)}}$ is a weighted sum of the previous memory cell $\mathbf{c_{B(A)}}$ and a function of the
current input.

\textbf{Proof of Symmetry}:
Because of the dual structure, it's straightforward to prove the symmetry of the proposed Loopy network.
If we use $\mathbf{F(x_A, x_B)}$ and $\mathbf{F(x_B, x_A)}$ denoting the final output hidden states
of different input orders $(\mathbf{x_A, x_B})$ and $(\mathbf{x_B, x_A})$ respectively, i.e., average of the first node's output. The symmetric property of matching requires that
$\mathbf{F(x_A, x_B)=F(x_B, x_A)}$.
As shown in Figure~\ref{fig2}, the hidden state of node $\mathbf{n_1}$ with input
$\mathbf{x_A}$ is denoted as $\mathbf{h_{n_1,A}}$. And $\mathbf{h_{n_1^{'},B}}$ is the hidden state of node $\mathbf{n_1^{'}}$ with input $\mathbf{x_B}$.
$\mathbf{F(x_A,x_B)}$ is the final hidden state which is used to determine the similarity with the input order $\mathbf{(x_A,x_B)}$. $\mathbf{F(x_A,x_B)}$ and $\mathbf{F(x_B,x_A)}$ are determined as follows:
\begin{equation}
\begin{aligned}
&\mathbf{F(x_A,x_B)}=(\mathbf{h_{n_1,A}+h_{n_1^{'},B}})/2,\\
&\mathbf{F(x_B,x_A)}=(\mathbf{h_{n_1,B}+h_{n_1^{'},A}})/2,\\
&\end{aligned}
\label{equf}
\vspace{-0.7cm}
\end{equation}

$LSTM$ and $LSTM^{'}$ share parameters, therefore
\begin{equation}
\begin{aligned}
&\mathbf{h_{n_1,A}}=\mathbf{h_{n_1^{'},A}},\\
&\mathbf{h_{n_1,B}}=\mathbf{h_{n_1^{'},B}},
\end{aligned}
\label{equh}
\end{equation}
from Equation~\ref{equf} and~\ref{equh},
\begin{equation}
\begin{aligned}
&\mathbf{F(x_A,x_B)}=\mathbf{F(x_B, x_A)}.
\end{aligned}
\end{equation}
Thus it's proven that the proposed Loopy RNN structure possess symmetry property.

\subsection{Loopy RNN for Image Matching}
To facilitate image (patch) matching, additional network components shall be augmented/modified to the basic structure of Loopy RNN. First, a feature extraction sub-network, i.e., a CNN network, is utilized to map the original image patch to a learned feature space, to input to the core structure of loopy RNN (denoted as FeatureNet in the rest of this paper). Second, as it is generally not feasible to train/test a loopy structure, we develop a simple yet effective structural approximation to convert a Loopy RNN into a conventional LSTM network to measure the similarity of a pair of features (denoted as MetricNet). Details are given as follows.

\textbf{FeatureNet}.
We adopt the feature network of MatchNet~\cite{han2015matchnet} as our network prototype for extracting deep features, named as FeatureNet. The structure of FeatureNet is modulated from AlexNet~\cite{krizhevsky2012imagenet}.
 The detailed network structure of FeatureNet is illustrated in Figure~\ref{fig3}. Note that the input patch size of FeatureNet is $64\times 64$ and the output feature dimension (which is connected to the core structure of Loopy RNN) is 4096. For the purpose of simplicity, in our work, dropout and local response normalization layers are omitted, since our input is image patch rather than the entire image (which is more complicated). Note that the feature extraction sub-networks for different input image patches share the same parameters.
{
\begin{figure*}[t]
\centering
\includegraphics[width=0.8\textwidth,trim=0 10 0 10]{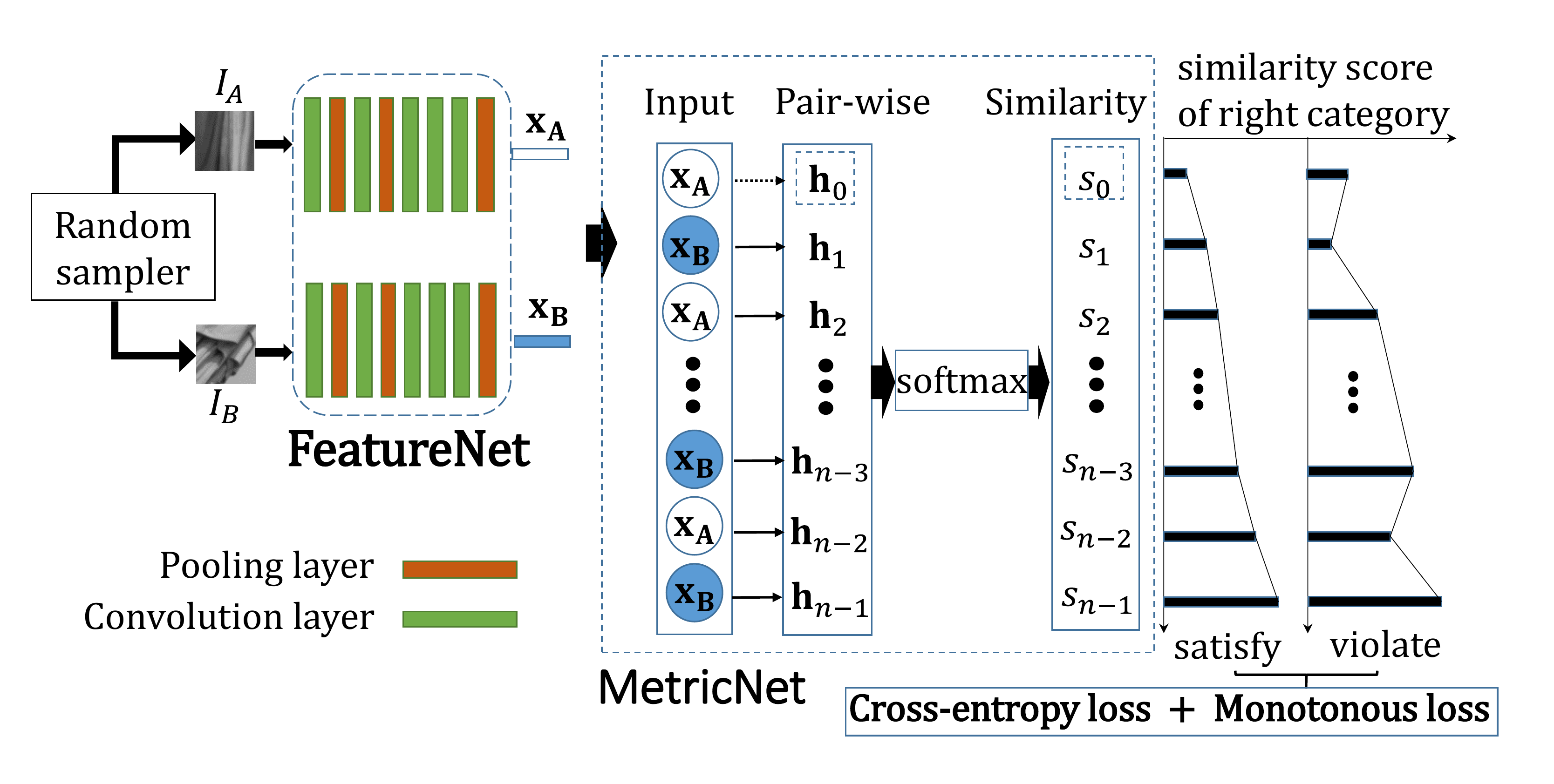}
\caption{The architecture of our network and the proposed monotonous loss. Right hand side shows two examples which satisfy (left) or violate (right) our monotonous cost.}
\label{fig3}
\vspace{-4mm}
\end{figure*}
}

\textbf{MetricNet}.
Although the two-node Loopy RNN has very simple structure, both forward and backward (training) computation for such a loopy structure is infeasible. Therefore, it is required to develop a simpler network structure which should not only mimic the recursive nature of Loopy RNN but also possess the advantage of easy training/testing. We observe that for a two-node Loopy RNN, the sequence of inputs could be regarded as an infinite repeat of both inputs. For example, for two image patches $I_A$ and $I_B$, their descriptors (the output feature vector from FeatureNet) are denoted by $\mathbf{x_{A}}$ and $\mathbf{x_{B}}$ respectively, then the input to the two-node Loopy RNN is the alternative sequence $\mathbf{x_{A}}\rightarrow \mathbf{x_{B}}\rightarrow \mathbf{x_{A}}\rightarrow...$. Motivated by this observation, we therefore utilize a conventional RNN/LSTM structure with a sequence of \textbf{finite} $K$ repeated $\mathbf{(x_A,x_B)}$ patterns as input to approximate the Loopy RNN matching network, as shown in Figure~\ref{fig3}. Thus it breaks the loopy structure to a standard RNN/LSTM structure, and off-the-shelf training algorithm could be directly applied. The proposed approximation has two advantages: 1) as information aggregates quickly through recursion, the output of network becomes stable after just a few repeats, i.e., $K$ could be small, which makes model training and inference very efficient; 2) if we regard each time step in the approximated LSTM network as an attention switch to one of the images, the proposed approximation naturally simulates the recursive matching process of human, i.e., iteratively switches attention between two input images and makes deeper and deeper comparisons.

Corresponding to the input sequence, our approximate LSTM network outputs a sequence of pair-wise matching features as $\mathbf{h}_{0}\rightarrow \mathbf{h}_{1}\rightarrow \mathbf{h}_{2}\rightarrow...\rightarrow \mathbf{h}_{n-3}\rightarrow \mathbf{h}_{n-2}\rightarrow \mathbf{h}_{n-1}$. Each pair-wise feature vector $\mathbf{h}_{n}$ encodes the similarity/comparison information aggregated from repeated input information from both images up to time step $n$. In contrast to previous matching algorithms which output only a scalar value to indicate similarity, our model outputs a feature vector to encode the similarity relationship between images, which conveys richer information and is more flexible for postprocessing (in cases where later fusion is preferred).
These output similarly feature vectors are sent to a softmax layer to determine the final similarity score, i.e., the final output of our network is a sequence of scores which describing the similarity of two image patches, denoted by $s_{0}\rightarrow s_{1}\rightarrow s_{2}\rightarrow...\rightarrow s_{n-3}\rightarrow s_{n-2}\rightarrow s_{n-1}$. $s_{n}$ is computed as follows:
\begin{equation}
\begin{aligned}
s_{n}=\frac{1}{1+e^{-\mathbf{\theta ^T} \mathbf{h}_{n}}}, n\neq 0,
\end{aligned}
\end{equation}
where $\mathbf{\theta}$ is the set of softmax layer parameters.
Each $s_{n}$ ranges in the interval $[0,1]$ and greater value indicates higher similarity. In fact, the first pair-wise feature $\mathbf{h_{0}}$ and first score $s_{0}$ are abandoned as Figure~\ref{fig3} showed, because only single patch information
instead of pair-wise information is included in first node.

\subsection{Monotonous Loss}
As the network's attention iteratively switches between two input image patches and more and more comparisons are made, the measurement for similarity should be more and more confident. In other words, it is required that the similarity score of the correct category should be monotonically non-decreasing as the information propagates deeper along the proposed matching network, as illustrated in Figure~\ref{fig3}.
However, a plain cross-entropy loss does not enforce such a monotonous non-decreasing property. We therefore use a monotonous loss~\cite{ma2016learning}, which extends the conventional cross-entropy loss to enforce the accuracy of prediction increase when the matching process goes deeper. Mathematically, we can express this loss as:
\begin{equation}
\begin{aligned}
&L_{n}^{m}=\max(0,(-1)^y(s_{n}-s_{n}^{pre})), n\neq 0,
\end{aligned}
\label{equloss}
\end{equation}
\begin{equation}
\begin{aligned}
&s_{n}^{pre}=
\begin{cases}
\max (s_{1}, s_{2} ..., s_{n-1}),& \text{$y=1$},\\
\min (s_{1}, s_{2} ..., s_{n-1}),& \text{$y=0$}.
\end{cases}
\end{aligned}
\end{equation}
Here, $L_{n}^{m}$ denotes the monotonous loss at time-step $n$, which penalizes the corresponding node if the output similarity score violates the monotonous
rule. $y$ is the ground truth label, i.e., $1$ for matched and $0$ for un-matched. $s_{n}$ is the predicted similarity score at time step $n$ and $s_{n}^{pre}$ is the
maximum ($y=1$) or minimum ($y=0$) prediction score until time step $n-1$. The max operation of Equation~\ref{equloss} picks out the nodes that violate the monotonous rule.
Denoted by $L_{n}^{c}$ as the standard cross-entropy loss, the overall loss could be expressed as:
\begin{equation}
\begin{aligned}
&L_{n}^{c}=-(y\log(s_n)+(1-y)\log(1-s_n)), n\neq 0,\\
&L_{n}=L_{n}^{c}+\lambda L_{n}^{m}, n\neq 0,
\end{aligned}
\end{equation}
where $\lambda$ is a weighting factor for both types of losses.


\subsection{Implementation Details}
\textbf{Data Preparation}.
Data imbalance is a common problem in patch-based image matching, because the number of positive pairs is far less than the number of negative pairs. A sampler which generates equal number of positive and negative pairs in a mini-batch is employed to prevent excessive bias to negative pairs.
To improve the generalization capability, we augment the dataset by vertically and horizontally flipping the original patch and rotating to $90$, $180$, $270$ degrees. Following the previous work~\cite{han2015matchnet}, we
map each pixel value $x$ (in [0,255]) to $(x-128)/160$.

\begin{table}[]
\centering\small
\begin{tabular}{c|c|c|c|c}
	\hline
\textbf{Name} & \textbf{Type} & \textbf{KS} & \textbf{S} & \textbf{OD} \\ \hline
Conv$0$ & C & 7x7 & 1 &$ 64\times64\times24$ \\
Pool$0$ & MP & 3x3 & 2 & $32\times32\times24$ \\
Conv$1$ & C & 5x5 & 1 & $32\times32\times64$ \\
Pool$1$ & MP & 3x3 & 2 & $16\times16\times64$ \\
Conv$2$ & C & 3x3 & 1 & $16\times16\times96$ \\
Conv$3$ & C & 3x3 & 1 & $16\times16\times96$ \\
Conv$4$ & C & 3x3 & 1 & $16\times16\times64$ \\
Pool$4$ & MP & 3x3 & 2 & $8\times8\times64 $\\
	\hline
\end{tabular}
\caption{Details of FeatureNet architecture. C: convolutional layer. MP: max pooling layer. KS: Kernel size. S: stride. OD: output dimension of feature map. OD is present as
(width $\times$ height $\times$ depth).}
\label{tab1}
\vspace{-0.4cm}
\end{table}
\textbf{Network Parameter and Training}.
The details of FeatureNet are listed in Table~\ref{tab1}.
For MetricNet, there are 3 key factors which influence the performance of Loopy RNN model: 1) the weighting factor
of monotonous loss $\lambda$; 2) the number of RNN nodes $N$ ($N \in \{6,8,10,12\}$); 3) the output dimension of LSTM node $D$ ($D \in \{512,1024,1536,2048\}$).

Our models are trained on Caffe~\cite{jia2014caffe} and optimized by Stochastic Gradient Descent (SGD) with the batch-size 32. Learning rate is set to 0.01 at the beginning and
decreased once every 1000 iterations. Our model converges to the steady state after about 70 epoches.

{
\begin{table*}[]
\centering\small
\begin{tabular}{c|c|c|c|c|c|c|c}
\hline
Train                              & \multicolumn{2}{c|}{Liberty}                                  & \multicolumn{2}{c|}{Notredame}                              & \multicolumn{2}{c|}{Yosemite}                                & \multicolumn{1}{l}{}     \\ \hline
Test                               & \multicolumn{1}{l|}{Notredame} & \multicolumn{1}{r|}{Yosemite} & \multicolumn{1}{l|}{Liberty} & \multicolumn{1}{l|}{Yosemite} & \multicolumn{1}{l|}{Liberty} & \multicolumn{1}{l|}{Notredame} & \multicolumn{1}{l}{mean} \\ \hline
nSIFT concat.+NNet                 & 14.35                         & 21.41                        & 20.44                       & 20.65                        & 22.23                       & 14.84                         & 18.99                    \\
MatchNet                           & 3.87                          & 10.88                        & 6.9                         & 8.39                         & 10.77                       & 5.67                          & 7.75                     \\
Siamese                            & 4.33                          & 14.89                        & 8.77                        & 13.23                        & 13.48                       & 5.75                          & 10.07                    \\
Pseudo-Siamese                     & 3.93                          & 12.5                         & 12.87                       & 12.64                        & 10.35                       & 5.44                          & 9.62                     \\
Siamese-2stream                    & 3.05                          & 9.02                         & 6.45                        & 10.44                        & 11.51                       & 5.29                          & 7.63                     \\
Siamese-2stream-$l_2$           & 4.54                          & 13.24                        & 8.79                        & 13.02                        & 12.84                       & 5.58                          & 9.67                     \\
2ch-2stream                     & \textbf{1.9}                     & \textbf{5}                            & \textbf{4.85}                        & \textbf{4.1}                        & \textbf{7.2}                  & \textbf{2.11}                          & \textbf{4.56}                    \\\hline
Loopy RNN(without monotonous loss) & 3.02                          & 8.92                         & 6.64                        & 8.13                         & 9.56                        & 3.96                          & 6.70                     \\
Loopy RNN(with monotonous loss)    & \textbf{2.79}                 & \textbf{8.29}                & \textbf{6.22}               & \textbf{7.71}                & \textbf{9.19}               & \textbf{3.72}                 & \textbf{6.32} \\ \hline
\end{tabular}
\caption{Matching result of UBC. We set our Loopy RNN model with $N=10$, $D=1024$ and $\lambda$ is set to 0.4 in the model with monotonous loss.}
\label{tab2}
\vspace{-5mm}
\end{table*}
}

\section{Experiments}

We evaluate our Loopy RNN network on two datasets: UBC dataset~\cite{winder2009picking} and Mikolajczyk Dataset~\cite{brown2011discriminative}. Extensive experimental evaluations
and in-depth analysis of the proposed method are presented in this section.
\subsection{Dataset and Evaluation Metric}

\textbf{UBC Dataset}. UBC includes three subsets: Liberty, Notredame and Yosemite. The number of image patches in the three subsets are 450k, 468k, and 634k respectively.
For the three subsets, 100k, 200k, and 500k pre-generated pairs are provided and the number of positive pairs equals to negative pairs.
Each patch in the dataset has a fixed size of $64\times 64$, which corresponds to the input dimension of our FeatureNet.

We follow the standard evaluation protocol~\cite{brown2011discriminative}, i.e., the model is iteratively trained on one subset and tested on the other two subsets,
FPR95 (false positive rate at $95\%$ recall) is adopted as the evaluation metric, the lower the better.

\textbf{Mikolajczyk Dataset}.
Mikolajczyk dataset is composed of 48 images in 8 sequences and each sequence corresponds to one of 5 transformations: viewpoint change, compression, blur, lighting change and zoom with
gradually increasing amount of transformation. Among 6 images of a sequence, one is reference image and the rest are transformed from reference image with different transformation magnitude,
i.e., the degree of transformation. The ground truth homography between reference image and transformation images are provided for evaluation.
We follow the method of~\cite{mikolajczyk2005performance} to test our model.
MAP (Mean Average Precision), which measures the area under the precision-recall curve is adopted as the evaluation metric.

\begin{figure}
\centering
\subfigure[Results of different $D$.]{
\includegraphics[width=0.22\textwidth, trim=0 20 0 20]{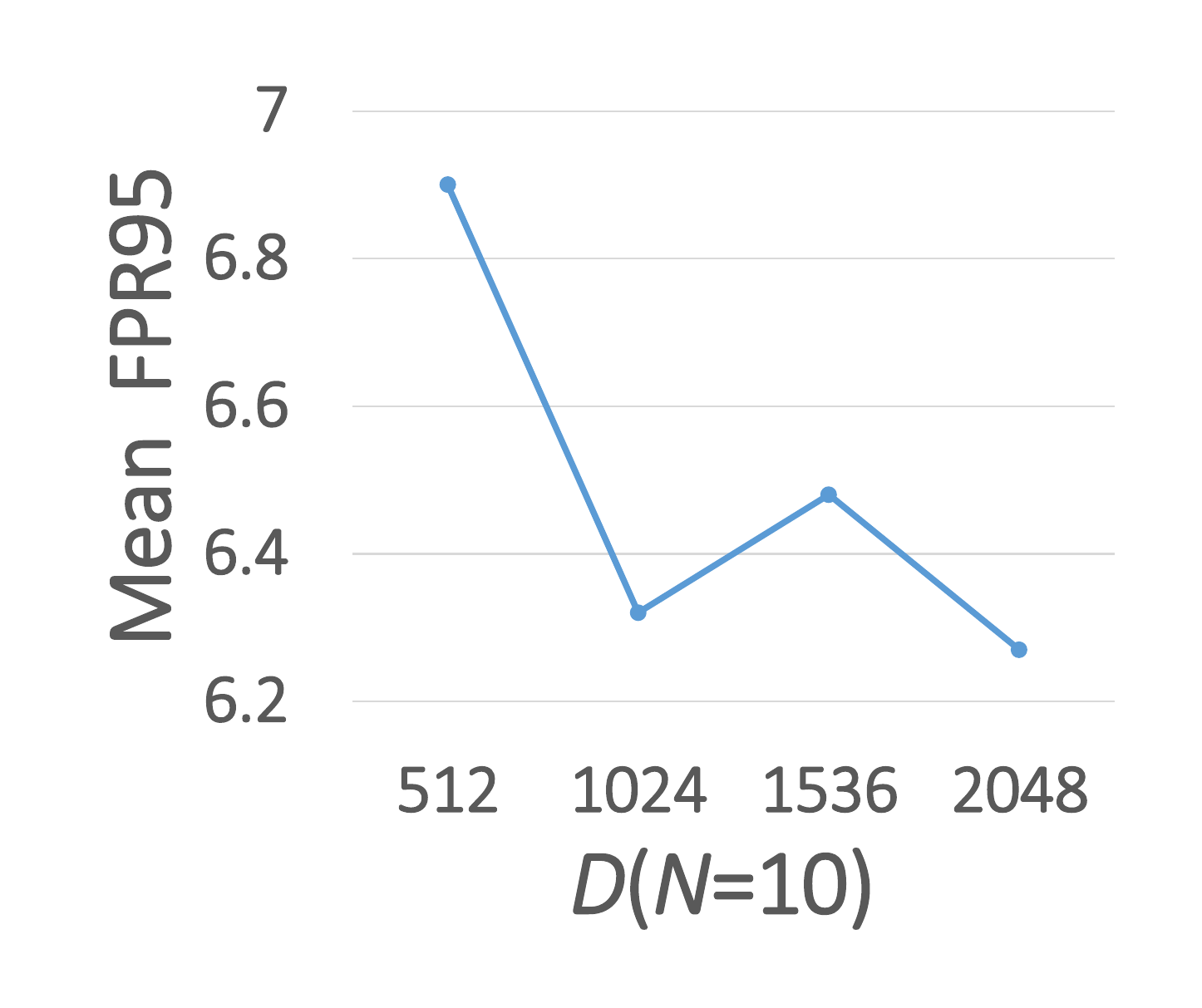}
}
\subfigure[Results of loopy time $N$.]{
\includegraphics[width=0.22\textwidth, trim=0 20 0 20]{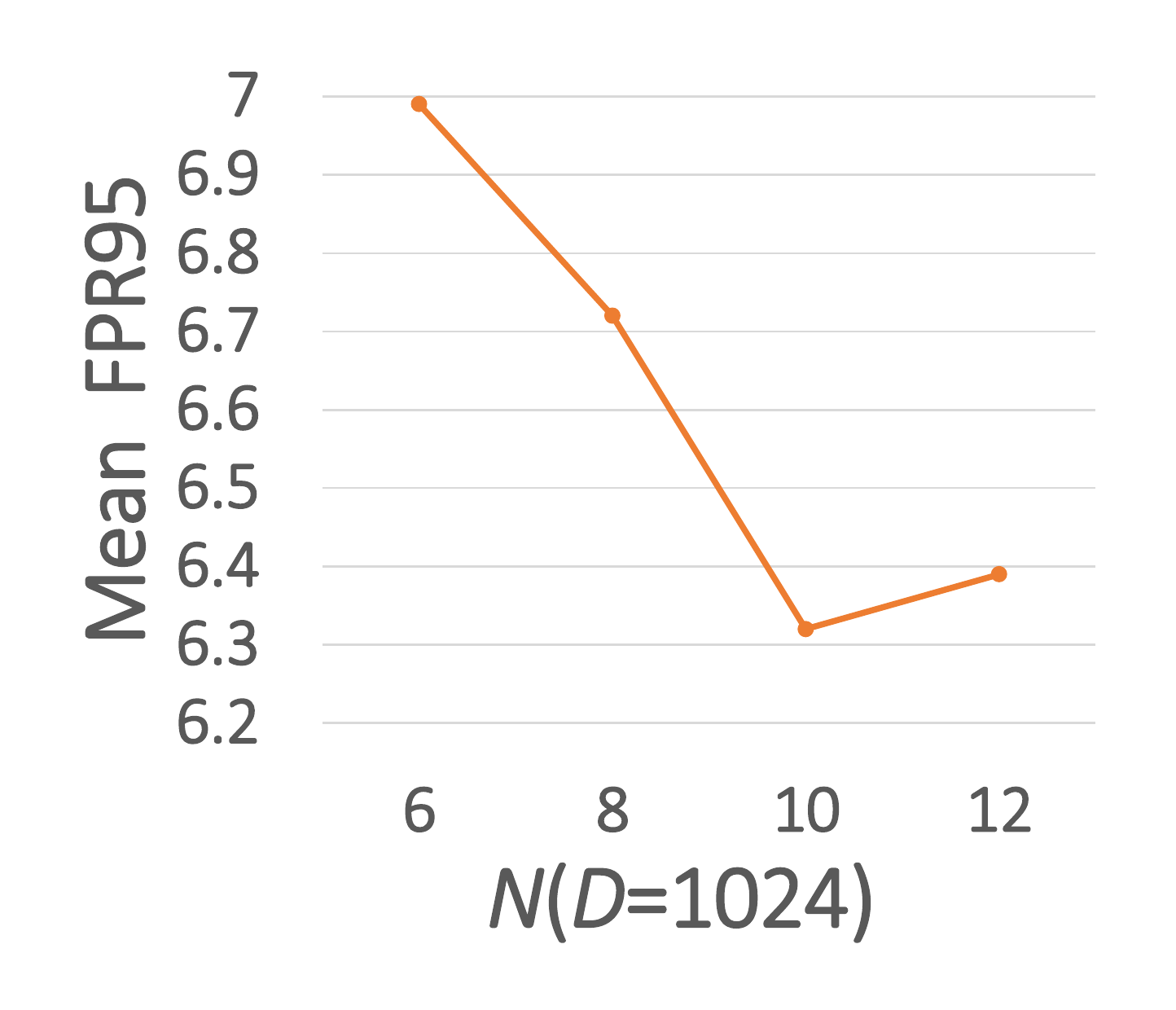}
}
\vspace{-5mm}
\caption{Results of different LSTM output dimension $D$ (with fixed $N=10$) and loopy time $N$ (with fixed $D=1024$).}
\label{figUBC}
\vspace{-3mm}
\end{figure}

\subsection{Evaluation on UBC}

In testing phase, we adopt the mean score of all nodes (except first node) as the final similarity score if monotonous loss is not used, i.e.,
$\lambda =0$. Otherwise, the final score is the mean of the last two nodes' score, i.e., for a sequence of 8 nodes, $s_{1}, s_{2}, ..., s_{7}$
are scores of last 7 nodes respectively, the final score $s=\frac{1}{7}(s_{1}+s_{2}+...+s_{7})$ if $\lambda =0$, otherwise $s=\frac{1}{2}(s_{6}+s_{7})$.

To verify the effectiveness of our Loopy RNN network, we compare our model with two recent works which apply CNN on patch-based image matching: MatchNet~\cite{han2015matchnet}
and ~\cite{zagoruyko2015learning}. Table~\ref{tab2} lists the comparison results. Our best model achieves $6.32\%$ average error rate with monotonous loss and parameter $N=10, D=1024$.
nSIFT concat.+NNet is the concatenation of SIFT feature and neural network. Our network outperforms this model by a large margin mainly because our feature extracted by FeatureNet
is more effective. Before metric network, MatchNet has the same architecture with our network and MatchNet achieves $7.75\%$ average FPR95. Therefore the 1.43\% promotion compared
with MatchNet completely comes from the combination of our Loopy RNN architecture and monotonous loss. The rest networks are all Siamese-like. Compared with Siamese and Pseudo-Siamese
network, the FPR95 of our network decreases 3.75\% and 3.3\% respectively. The gain comes from the FeatureNet and our Loopy RNN network. Siamese-2stream and Siamese-2stream-$l_2$
utilize information of different resolutions and achieves $7.63\%$ and $9.67\%$ average FPR95 respectively. Even only using information of one resolution, our model still obtains 1.31\%, 3.34\% improvement on FPR95.
To verify the effect of monotonous loss, we list the results of our model without monotonous loss. It is obvious that the performance decreases compared with the model with monotonous loss.
The experiment results illustrate that our Loopy RNN architecture has superiority over Siamese network and monotonous loss assists Loopy RNN to obtain further performance gain.

2ch-2stream network~\cite{zagoruyko2015learning} obtains better performance than our network. It achieves the performance by treating two grayscale
patches as two channels of a new image and classifying the new image into two category. The 2ch-2stream network performs better because it disposes two images jointly at the very beginning. Then each feature map includes the pair's feature. Because feature map can't input to the LSTM node directly, in our network, it is necessary to extract feature vector from image. Thus our FeatureNet disposes two images respectively. Only in MetricNet, pair's feature is disposed. From image to feature vector, our network misses some pair information compared with the 2-channel network.

\textbf{Parameter Analysis.}
On one hand, we find that large $\lambda$ makes the proposed network hard to converge. Thus we can not set $\lambda$ to a large value. On the other hand, $\lambda$ is used for balancing the weights of monotonous loss and cross-entropy loss. Too small value of $\lambda$ weakens the function of monotonous loss. As a result, we set $\lambda=0.4$ empirically. Figure~\ref{figUBC} shows the experiment results with different LSTM output dimension $D$ and loopy time $N$. We fix $N$ as 10 and test the model with different $D$ (Figure~\ref{figUBC}(a)), then we fix $D$ as 1024 and test our model with different $N$ (Figure~\ref{figUBC}(b)). We observe that higher dimension results to better performance as well as computational complexity. Here, we set $D$ as 1024. For $N$, larger loopy time promotes the performance, however, when $N$ exceeds 10, the performance saturates. Thus we set $N$ to 10. This is because that the network already has enough observations for making the judgement. Based on the above analysis, we choose the model with parameter $N=10$, $D=1024$ as our best models even though the $N=10$, $D=2048$ model outperforms
the former a little.

\subsection{Evaluation On Local Descriptors}

{
\begin{figure}[]
\centering
\includegraphics[width=0.44\textwidth,trim=0 60 0 20]{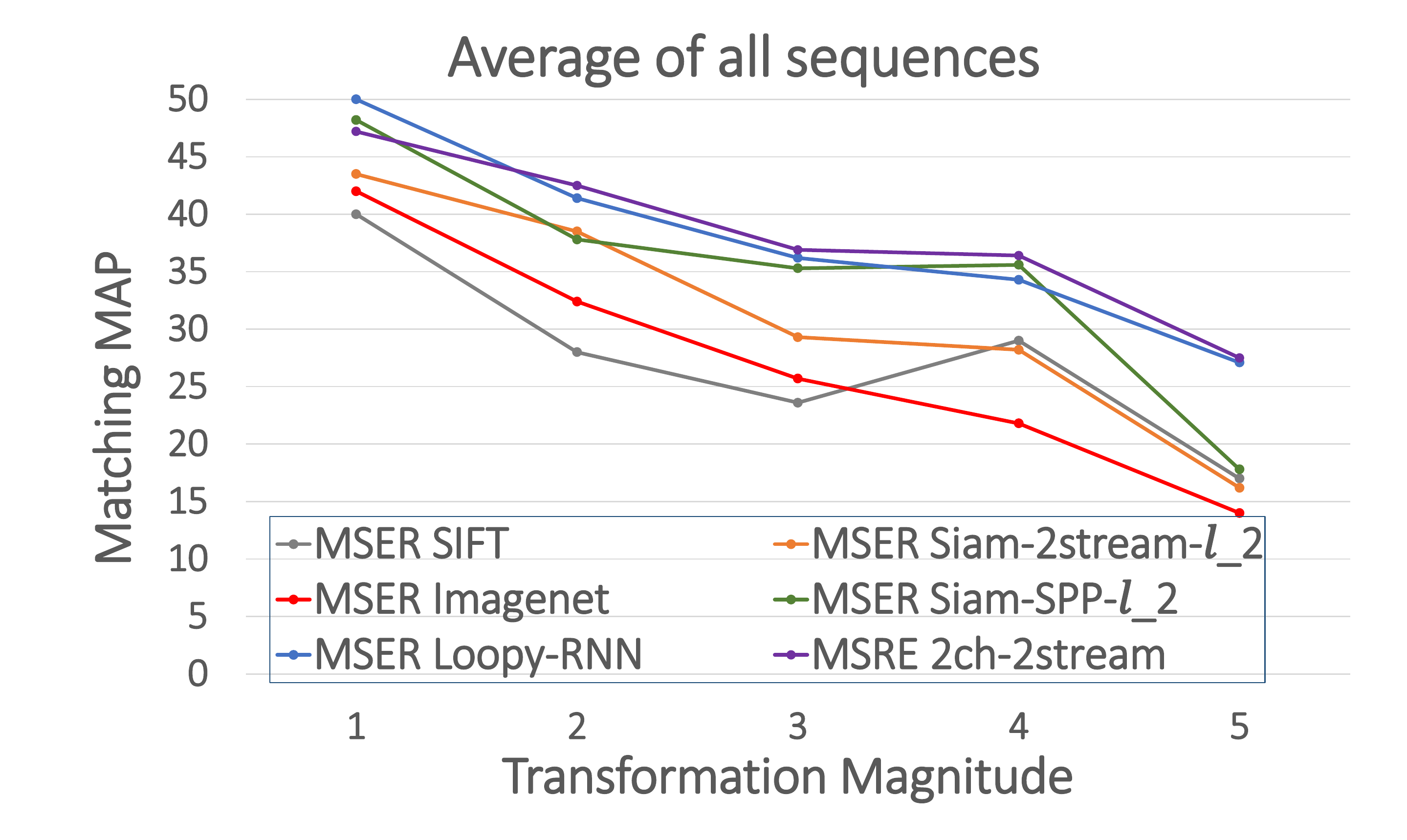}
\caption{Evaluation on the Mikolajczyk dataset. Our model is trained on Liberty with $N=10$, $D=1024$.}
\label{figell}
\vspace{-5mm}
\end{figure}
}

We compare our network with five networks of~\cite{zagoruyko2015learning} that are tested on the same dataset. MSER SIFT uses the Euclidean distance
of SIFT features to measure the similarity of two patches. The other three models, MSER Imagenet, MSER Siam-2stream-$l_2$ and MSER Siam-SPP-$l_2$ substitute SIFT feature with CNN feature. MSER 2ch-2stream dispose 2 images as mentioned above. All the models are trained on the Liberty dataset.
We test our model under different transformation with increasing magnitudes. Figure 5 illustrates the overall
results. Our network outperforms most networks, especially in the extreme case. When transformation magnitude equals to 5, our network and 2ch-2stream greatly outperforms other networks by approximately 10\%,
which demonstrates the robustness of our model. MSER Siam-SPP-$l_2$ is a network with SPP layer
which makes the network able to deal with patches of different scale. Although our model only deals with single scale patches, it still outperforms MSER Siam-SPP-$l_2$ in most cases. When magnitude equals to 1, 2, 5, MAP of our model achieves better performances by about
2\%, 4\% and 9\% respectively compared with MSER Siam-SPP-$l_2$.
The performance of our network is comparable with the 2ch-2stream network. This indicates that our loopy network has good generalization ability as it performs well on different datasets.
\vspace{-4mm}
\section{Conclusion}

In this paper, we propose a novel Loopy RNN which matches a pair of patches in a recurrent manner. Based on widely used cross-entropy loss, we add the
monotonous loss aiming at restricting the output of a sequence. Combined with monotonous cross-entropy loss, our
network imitates human to observe the two patches back and forth and the judgement become more and more confident in this process. Our experimental results show the effectiveness of the proposed method.

\vspace{-4mm}
\section*{Acknowledgements}\label{sec:acknow}
{The work was supported by State Key Research and Development Program (2016YFB1001003).
This work was partly supported by NSFC (61502301), China's Thousand Youth Talents Plan, National Natural Science Foundation of China (61521062), the 111 Project (B07022) and the Opening Project of Shanghai Key Laboratory of Digital Media Processing and Transmissions.}

\bibliographystyle{named}
\bibliography{image_matching_via_loopy_RNN}

\begin{thebibliography}{}

\bibitem[\protect\citeauthoryear{Arandjelovic \bgroup \em et al.\egroup
  }{2016}]{arandjelovic2016netvlad}
Relja Arandjelovic, Petr Gronat, Akihiko Torii, Tomas Pajdla, and Josef Sivic.
\newblock Netvlad: Cnn architecture for weakly supervised place recognition.
\newblock In {\em CVPR}, 2016.

\bibitem[\protect\citeauthoryear{Bay \bgroup \em et al.\egroup
  }{2006}]{bay2006surf}
Herbert Bay, Tinne Tuytelaars, and Luc~Van Gool.
\newblock Surf: Speeded up robust features.
\newblock In {\em ECCV}, 2006.

\bibitem[\protect\citeauthoryear{Bertinetto \bgroup \em et al.\egroup
  }{2016}]{bertinetto2016fully}
Luca Bertinetto, Jack Valmadre, Jo{\~a}o~F Henriques, Andrea Vedaldi, and
  Philip~HS Torr.
\newblock Fully-convolutional siamese networks for object tracking.
\newblock In {\em ECCV}, 2016.

\bibitem[\protect\citeauthoryear{Bromley \bgroup \em et al.\egroup
  }{1993}]{bromley1993signature}
Jane Bromley, James~W Bentz, Bottou L{\'e}on, Isabelle Guyon, Yann LeCun, Cliff
  Moore, Eduard S{\"a}ckinger, and Roopak Shah.
\newblock Signature verification using a ��siamese�� time delay neural
  network.
\newblock {\em IJPRAI}, pages 669--688, 1993.

\bibitem[\protect\citeauthoryear{Brown \bgroup \em et al.\egroup
  }{2011}]{brown2011discriminative}
Matthew Brown, Gang Hua, and Simon Winder.
\newblock Discriminative learning of local image descriptors.
\newblock {\em IEEE TPAMI}, pages 43--57, 2011.

\bibitem[\protect\citeauthoryear{Cheng \bgroup \em et al.\egroup
  }{2014}]{cheng2014fast}
Jian Cheng, Cong Leng, Jiaxiang Wu, Hainan Cui, and Hanqing Lu.
\newblock Fast and accurate image matching with cascade hashing for 3d
  reconstruction.
\newblock In {\em CVPR}, 2014.

\bibitem[\protect\citeauthoryear{Dorffner}{1996}]{dorffner1996neural}
Georg Dorffner.
\newblock Neural networks for time series processing.
\newblock In {\em Neural network world}, 1996.

\bibitem[\protect\citeauthoryear{Fischer \bgroup \em et al.\egroup
  }{2014}]{fischer2014descriptor}
Philipp Fischer, Alexey Dosovitskiy, and Thomas Brox.
\newblock Descriptor matching with convolutional neural networks: a comparison
  to sift.
\newblock {\em arXiv preprint arXiv:1405.5769}, 2014.

\bibitem[\protect\citeauthoryear{Han \bgroup \em et al.\egroup
  }{2015}]{han2015matchnet}
Xufeng Han, Thomas Leung, Yangqing Jia, Rahul Sukthankar, and Alexander~C Berg.
\newblock Matchnet: Unifying feature and metric learning for patch-based
  matching.
\newblock In {\em CVPR}, 2015.

\bibitem[\protect\citeauthoryear{Hochreiter and
  Schmidhuber}{1997}]{hochreiter1997long}
Sepp Hochreiter and J{\"u}rgen Schmidhuber.
\newblock Long short-term memory.
\newblock {\em Neural computation}, pages 1735--1780, 1997.

\bibitem[\protect\citeauthoryear{Jain \bgroup \em et al.\egroup
  }{2012}]{jain2012metric}
Prateek Jain, Brian Kulis, Jason~V Davis, and Inderjit~S Dhillon.
\newblock Metric and kernel learning using a linear transformation.
\newblock {\em JMLR}, pages 519--547, 2012.

\bibitem[\protect\citeauthoryear{Jia and Darrell}{2011}]{jia2011heavy}
Yangqing Jia and Trevor Darrell.
\newblock Heavy-tailed distances for gradient based image descriptors.
\newblock In {\em NIPS}, 2011.

\bibitem[\protect\citeauthoryear{Jia \bgroup \em et al.\egroup
  }{2014}]{jia2014caffe}
Yangqing Jia, Evan Shelhamer, Jeff Donahue, Sergey Karayev, Jonathan Long, Ross
  Girshick, Sergio Guadarrama, and Trevor Darrell.
\newblock Caffe: Convolutional architecture for fast feature embedding.
\newblock In {\em ACM international conference on Multimedia}, 2014.

\bibitem[\protect\citeauthoryear{Krizhevsky \bgroup \em et al.\egroup
  }{2012}]{krizhevsky2012imagenet}
Alex Krizhevsky, Ilya Sutskever, and Geoffrey~E Hinton.
\newblock Imagenet classification with deep convolutional neural networks.
\newblock In {\em NIPS}, 2012.

\bibitem[\protect\citeauthoryear{Lowe}{2004}]{lowe2004distinctive}
David~G Lowe.
\newblock Distinctive image features from scale-invariant keypoints.
\newblock {\em IJCV}, pages 91--110, 2004.

\bibitem[\protect\citeauthoryear{Ma \bgroup \em et al.\egroup
  }{2016}]{ma2016learning}
Shugao Ma, Leonid Sigal, and Stan Sclaroff.
\newblock Learning activity progression in lstms for activity detection and
  early detection.
\newblock In {\em Proceedings of the IEEE Conference on Computer Vision and
  Pattern Recognition}, 2016.

\bibitem[\protect\citeauthoryear{Mikolajczyk and
  Schmid}{2005}]{mikolajczyk2005performance}
Krystian Mikolajczyk and Cordelia Schmid.
\newblock A performance evaluation of local descriptors.
\newblock {\em IEEE TPAMI}, 27(10):1615--1630, 2005.

\bibitem[\protect\citeauthoryear{Paulin \bgroup \em et al.\egroup
  }{2015}]{paulin2015local}
Mattis Paulin, Matthijs Douze, Zaid Harchaoui, Julien Mairal, Florent Perronin,
  and Cordelia Schmid.
\newblock Local convolutional features with unsupervised training for image
  retrieval.
\newblock In {\em CVPR}, 2015.

\bibitem[\protect\citeauthoryear{Ren \bgroup \em et al.\egroup
  }{2017}]{ren2017unsupervised}
Zhe Ren, Junchi Yan, Bingbing Ni, Bin Liu, Xiaokang Yang, and Hongyuan Zha.
\newblock Unsupervised deep learning for optical flow estimation.
\newblock In {\em Thirty-First AAAI Conference on Artificial Intelligence},
  2017.

\bibitem[\protect\citeauthoryear{Rublee \bgroup \em et al.\egroup
  }{2011}]{rublee2011orb}
Ethan Rublee, Vincent Rabaud, Kurt Konolige, and Gary Bradski.
\newblock Orb: An efficient alternative to sift or surf.
\newblock In {\em ICCV}, 2011.

\bibitem[\protect\citeauthoryear{Shyam \bgroup \em et al.\egroup
  }{2017}]{shyam2017attentive}
Pranav Shyam, Shubham Gupta, and Amdebkar Dukkipati.
\newblock Attentive recurrent comparators.
\newblock {\em arXiv preprint arXiv:1703.00767}, 2017.

\bibitem[\protect\citeauthoryear{Tola \bgroup \em et al.\egroup
  }{2008}]{tola2008fast}
Engin Tola, Vincent Lepetit, and Pascal Fua.
\newblock A fast local descriptor for dense matching.
\newblock In {\em CVPR}, 2008.

\bibitem[\protect\citeauthoryear{Trzcinski \bgroup \em et al.\egroup
  }{2012}]{trzcinski2012learning}
Tomasz Trzcinski, Mario Christoudias, Vincent Lepetit, and Pascal Fua.
\newblock Learning image descriptors with the boosting-trick.
\newblock In {\em NIPS}, 2012.

\bibitem[\protect\citeauthoryear{Winder \bgroup \em et al.\egroup
  }{2009}]{winder2009picking}
Simon Winder, Gang Hua, and Matthew Brown.
\newblock Picking the best daisy.
\newblock In {\em CVPR}, 2009.

\bibitem[\protect\citeauthoryear{Yan \bgroup \em et al.\egroup
  }{2015a}]{yan2015consistency}
Junchi Yan, Jun Wang, Hongyuan Zha, Xiaokang Yang, and Stephen Chu.
\newblock Consistency-driven alternating optimization for multigraph matching:
  A unified approach.
\newblock {\em IEEE Transactions on Image Processing}, 24(3):994--1009, 2015.

\bibitem[\protect\citeauthoryear{Yan \bgroup \em et al.\egroup
  }{2015b}]{yan2015discrete}
Junchi Yan, Chao Zhang, Hongyuan Zha, Wei Liu, Xiaokang Yang, and Stephen~M
  Chu.
\newblock Discrete hyper-graph matching.
\newblock In {\em Proceedings of the IEEE Conference on Computer Vision and
  Pattern Recognition}, 2015.

\bibitem[\protect\citeauthoryear{Yan \bgroup \em et al.\egroup
  }{2016a}]{yan2016multi}
Junchi Yan, Minsu Cho, Hongyuan Zha, Xiaokang Yang, and Stephen~M Chu.
\newblock Multi-graph matching via affinity optimization with graduated
  consistency regularization.
\newblock {\em IEEE transactions on pattern analysis and machine intelligence},
  38(6):1228--1242, 2016.

\bibitem[\protect\citeauthoryear{Yan \bgroup \em et al.\egroup
  }{2016b}]{yan2016person}
Yichao Yan, Bingbing Ni, Zhichao Song, Chao Ma, Yan Yan, and Xiaokang Yang.
\newblock Person re-identification via recurrent feature aggregation.
\newblock In {\em ECCV}, 2016.

\bibitem[\protect\citeauthoryear{Yi \bgroup \em et al.\egroup
  }{2014}]{yi2014deep}
Dong Yi, Zhen Lei, Shengcai Liao, Stan~Z Li, et~al.
\newblock Deep metric learning for person re-identification.
\newblock In {\em ICPR}, 2014.

\bibitem[\protect\citeauthoryear{Zagoruyko and
  Komodakis}{2015}]{zagoruyko2015learning}
Sergey Zagoruyko and Nikos Komodakis.
\newblock Learning to compare image patches via convolutional neural networks.
\newblock In {\em CVPR}, 2015.

\end{thebibliography}

\end{document}